%% file: 0_main.tex
\documentclass[a4paper,fleqn]{cas-dc}

\usepackage[authoryear]{natbib}

\usepackage{xcolor}

\usepackage[linesnumbered,ruled,lined,boxed]{algorithm2e}

\usepackage{caption}
\usepackage{subcaption}

\usepackage{amsmath}
\usepackage{amssymb}
\usepackage{amsthm}

\usepackage{enumitem}
\usepackage{tikz}
\usetikzlibrary{patterns}
\usetikzlibrary{positioning}
\usetikzlibrary{decorations.pathreplacing,angles,quotes}

\usepackage[utf8]{inputenc}
\usepackage[T1]{fontenc}

\usepackage{dirtytalk}

\makeatletter
\def\ps@pprintTitle{%
 \let@oddhead@empty
 \let@evenhead@empty
 \def@oddfoot{\centerline{\thepage}}%
 \let@evenfoot@oddfoot}
\makeatother

\begin{document}
\let\WriteBookmarks\relax
\def\floatpagepagefraction{1}
\def\textpagefraction{.001}
\shorttitle{An anytime tree search for the 2018 ROADEF/EURO challenge}
\shortauthors{Libralesso, Fontan}

\title[mode = title]{An anytime tree search algorithm for the 2018 ROADEF/EURO challenge glass cutting problem}                      
\author[1,2]{Luc Libralesso}
\ead{luc.libralesso@grenoble-inp.fr}

\author[1,2]{Florian Fontan}
\ead{dev@florian-fontan.fr}

\address[1]{Univ. Grenoble Alpes, CNRS, Grenoble INP, G-SCOP, 38000 Grenoble, France}
\address[2]{Each author contributes equally to this work.}

\begin{abstract}
In this article, we present the anytime tree search algorithm we designed for the 2018 ROADEF/EURO challenge glass cutting problem proposed by the French company Saint-Gobain. The resulting program was ranked first among 64 participants. Its key components are: a new search algorithm called Memory Bounded A* (MBA*) with guide functions, a symmetry breaking strategy, and a pseudo-dominance rule. We perform a comprehensive study of these components showing that each of them contributes to the algorithm global performances.
In addition, we designed a second tree search algorithm fully based on the pseudo-dominance rule and dedicated to some of the challenge instances with strong precedence constraints. On these instances, it finds the best-known solutions very quickly.
\end{abstract}

\begin{highlights}
\item We present the algorithm that allowed us to win the final phase of the 2018 ROADEF/EURO challenge.
\item We introduce a new generic anytime tree search algorithm called \emph{Memory Bounded A*}.
\item We outline guide functions, a symmetry breaking strategy, and a pseudo-dominance rule.
\item We show by a comprehensive study that each of these components contributes to the algorithm global performance.
\item We describe another tree search algorithm fully based on the pseudo-dominance rule and dedicated to instances with strong precedence constraints, for which it quickly finds the best-known solutions
\end{highlights}


\maketitle

\section{Introduction}
In automated planning and scheduling (AI planning) communities, resolution methods often involve exploring a search tree. These methods usually perform an advanced greedy procedure. They explore the search tree starting with the parts that are evaluated \textit{a priori} most promising and continue the exploration (so long as computational time is provided). Such methods are called \emph{anytime tree searches} because they can be stopped at any time and provide good solutions relatively to the allowed computation time. Thus, they share the same purpose as classical meta-heuristics, but are less common in Operations Research. Still, one called Beam Search has shown a relative popularity in the \emph{Cutting \& Packing} literature \citep{akeb2009beam,bennell2010beam,baldi2014branch}. They usually perform little inference within each node, and thus can open millions of nodes per second.

On the other hand, branch \& bound algorithms are ubiquitous in Operations Research. Such methods are usually designed to prove optimality, thus relying on strong bound computations (such as Lagrangian relaxations) and advanced pruning rules (dominances, symmetries, etc.)\ to reduce the size of the search tree as much as possible. However, such refined computations often drastically reduce the number of nodes opened per second. Consequently, the quality of solutions obtained on larger instances can be harmed due to fewer nodes opened.

The constructive nature of both anytime tree search algorithms and branch \& bounds suggests that it could be possible to incorporate their respective advantages in a common approach. Indeed, some branch \& bound components may be (relatively) inexpensive to compute while still greatly reducing the search space. Using them within an anytime tree search algorithm would allow getting the best of both methods. It would provide a constructive method that is designed to find good solutions fast while taking advantage of the search space reductions from branch \& bounds.

With this in mind, we decided to develop such an algorithm for the 2018 ROADEF/EURO challenge glass cutting problem.
We may note that this is unusual as almost all top-ranked methods in previous editions of the challenge mainly rely on local search or mathematical programming techniques.

\bigskip

We propose an anytime tree search with some simple bounds, pseudo-dominance properties, and symmetry breaking rules.
We introduce some new guidance strategy that allows the algorithm to perform significantly better than if it was guided by a bound as in classical branch \& bound methods.

The search strategy can be roughly described as follows. It is a restarting strategy that starts its first iteration by performing very aggressive heuristic prunings.
At the second iteration, it performs less aggressive heuristic prunings, taking more time than the previous iteration, but finding better solutions. 
If the algorithm runs long enough, some iteration may perform no heuristic pruning, thus the method will be able to guarantee optimality. The resulting method obtained the best results compared to the other submitted approaches during the final phase. We named it \emph{Memory Bounded A*} as it performs a series of A* with heuristic prunings which guarantee no-more than a given amount of nodes active at the same time.

We also highlight a general methodology that can be applied to other complex problems (and with other tree search algorithms). 
Indeed, the method can be divided into two parts: the \emph{Branching Scheme}, usually problem-specific, which is a definition of the
implicit search tree (\emph{i.e.}\  root node, how to generate children of a given node, lower bounds, dominance rules, \emph{etc.}); and a strategy, usually generic, to explore the tree.
This decomposition allows rapid prototyping of both search tree definitions and tree search algorithms as many generic parts can be reused within other algorithms. It also helps to draw insights about the contribution of each component to the resulting search algorithm.

%This paper presents the algorithm we designed to tackle the challenge. The challenge generalizes many problems that can be found in the \emph{Cutting \& Packing} literature. A natural question would be to evaluate how the presented algorithm performs on these variants. We wrote a companion paper {\color{red}fontan2020} answering this question (it appears that it is competitive with the state-of-the-art on many variants).
%Thus, the present paper mainly focuses on algorithmic ideas and the challenge specific parts (namely the min-waste constraint and \emph{Dynamic Programming A*}). We refer the reader to {\color{red}fontan2020} for more information about how this algorithm generalizes to other guillotine \emph{Cutting \& Packing} problems.
%We also believe that many tree search techniques from AI can be applied with success on many Operations Research problems.

This paper is structured as follows.
In Section~\ref{sec:chal:pb}, we state the problem constraints and objective. 
In Section~\ref{sec:chal:defs}, we give some notations and definitions.
In Section~\ref{sec:chal:bs}, we describe the branching scheme and in Section \ref{sec:chal:ts}, the tree search algorithm we designed.
Finally, in Section \ref{sec:chal:num}, we show the numerical results we obtained.

\section{Problem description}
\label{sec:chal:pb}
\input{1_pb_description}

\section{Definitions and notations}
\label{sec:chal:defs}
\input{2_definitions}

\section{Branching scheme}
\label{sec:chal:bs}
\input{2_branching_scheme}

\section{Tree search}
\label{sec:chal:ts}
\input{3_tree_search}

\section{Numerical results}
\label{sec:chal:num}
\input{4_num_results}

\section{Conclusion and perspectives}
\label{sec:conc}
\input{5_conclusion}

\bibliographystyle{cas-model2-names}
\bibliography{0_main}

\end{document}

%% file: 1_pb_description.tex
The 2018 ROADEF/EURO challenge was dedicated to an industrial cutting problem from the French company \emph{Saint-Gobain}. The challenge consists in packing rectangular glass items into standardized bins of dimensions $W \times H$ (6m $\times$ 3.21m).

\begin{figure}[]
  \centering
  \begin{tikzpicture}
    \draw[fill=gray!30] (0,0) rectangle (6,3.21);
    \draw[fill=white] (0,0) rectangle (2.7,2) node[pos=.5] {$J_1$};
    \draw[dotted] (2.8,2) -- (3,2);
    \draw[fill=white] (0,2) rectangle (2.8,3.21) node[pos=.5] {$J_2$};
    \draw[fill=white] (3,0) rectangle (5,3.21) node[pos=.5] {$J_3$};
    \draw[fill=gray!60] (5,0) rectangle (6,3.21);
  \end{tikzpicture}
  \caption{Optimal solution of the case containing the following three items with the chain precedence constraint $J_1 \to J_2 \to J_3$. Additional waste must be added before the first $1$-cut. Otherwise either the waste area to the right of $J_1$ or the waste area to the right of $J_2$ would violate the minimum waste constraint.}
  \label{fig:mincuts}
\end{figure}
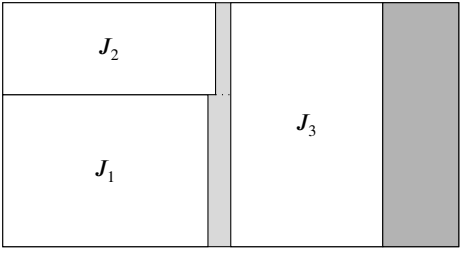

The cutting plan needs to satisfy the following constraints:
\begin{itemize}
  \item All items need to be produced
  \item Item rotation is allowed
  \item Cuts must be of guillotine type. Figure~\ref{fig:guillotine_illustration} illustrates two examples of non-guillotine and guillotine patterns. Furthermore, the number of stages is limited to four, with only one $4$-cut allowed on a sub-plate obtained after $3$-cuts. This configuration is close to classical three-staged non-exact guillotine patterns, but differs in that a sub-plate obtained after $3$-cuts may contain two items as illustrated in Figure~\ref{tikz:4cuts}.

\begin{figure}[]
  \begin{subfigure}[t]{0.5\textwidth}
    \centering
    \includegraphics[width=4.5cm]{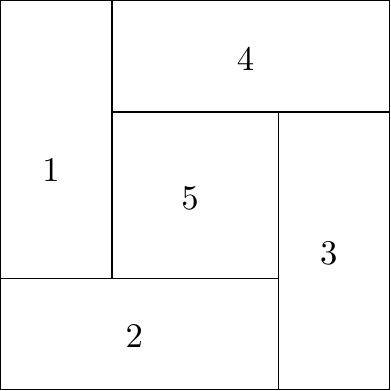}
    \caption{Non-guillotine pattern}
  \end{subfigure}

  \medskip

  \begin{subfigure}[t]{0.5\textwidth}
    \centering
    \includegraphics[width=4.5cm]{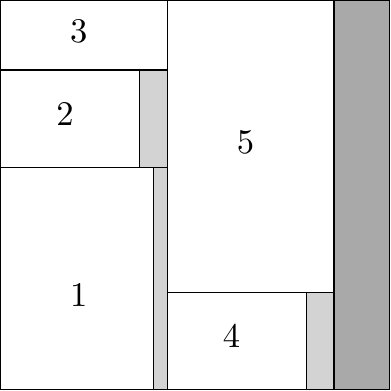}
    \caption{Guillotine pattern}
  \end{subfigure}
  \caption{Illustration of a non-guillotine pattern (a) and a guillotine one (b)}
  \label{fig:guillotine_illustration}
\end{figure}

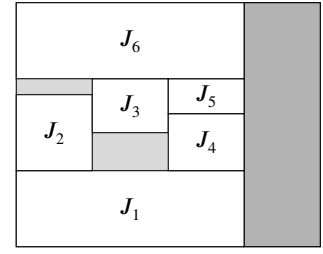
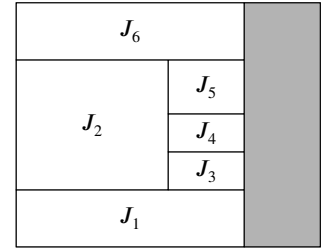
\begin{figure}[]
  \centering
  \begin{subfigure}[t]{0.5\textwidth}
    \centering
    \begin{tikzpicture}
      \draw[fill=gray!30] (0,0) rectangle (4,3.21);
      \draw[fill=white] (0,0) rectangle (3,1) node[pos=.5] {$J_1$};
      \draw[fill=white] (0,1) rectangle (1,2) node[pos=.5] {$J_2$};
      \draw[fill=white] (1,1.5) rectangle (2,2.21) node[pos=.5] {$J_3$};
      \draw[fill=white] (2,1) rectangle (3,1.75) node[pos=.5] {$J_4$};
      \draw[fill=white] (2,1.75) rectangle (3,2.21) node[pos=.5] {$J_5$};
      \draw[fill=white] (0,2.21) rectangle (3,3.21) node[pos=.5] {$J_6$};
      \draw[fill=gray!60] (3,0) rectangle (4,3.21);
    \end{tikzpicture}
    \caption{}
  \end{subfigure}

  \medskip

  \begin{subfigure}[t]{0.5\textwidth}
    \centering
    \begin{tikzpicture}
      \draw[fill=gray!30] (0,0) rectangle (4,3.21);
      \draw[fill=white] (0,0) rectangle (3,0.75) node[pos=.5] {$J_1$};
      \draw[fill=white] (0,0.75) rectangle (2,2.46) node[pos=.5] {$J_2$};
      \draw[fill=white] (2,0.75) rectangle (3,1.25) node[pos=.5] {$J_3$};
      \draw[fill=white] (2,1.25) rectangle (3,1.75) node[pos=.5] {$J_4$};
      \draw[fill=white] (2,1.75) rectangle (3,2.46) node[pos=.5] {$J_5$};
      \draw[fill=white] (0,2.46) rectangle (3,3.21) node[pos=.5] {$J_6$};
      \draw[fill=gray!60] (3,0) rectangle (4,3.21);
    \end{tikzpicture}
    \caption{}
  \end{subfigure}

  \caption{Only one $4$-cut is allowed. Therefore, pattern (a) is feasible but pattern (b) is not}
  \label{tikz:4cuts}
\end{figure}

 \item Items are subject to chain precedence constraints. The extraction order is as follows:
   rightmost first level sub-plates first;
   within a first level sub-plate, bottommost second level sub-plates first;
   within a second level sub-plate, rightmost items first;
   and within a third level sub-plate, bottommost item first.
   Most instances have a dozen chains, three instances have $2$ chains and five instances are not subject to precedence constraints.
 \item Bins contain defects (between $0$ and $8$ rectangles about a few centimeters high and wide). Items must be defect-free and it is forbidden to cut through a defect. Even if the bins have the same dimensions, the presence of defects makes the set of bins heterogenous. It is important to note that bins must be used in the order they are given.
 \item Depending on their level, sub-plates are subject to minimum and maximum size constraints. The width of first level sub-plates must lie between $w_\text{min}^1 = 100$ and $w_\text{max}^1 = 3500$, except for wastes. The height of second-level sub-plates must be at least $w_\text{min}^2 = 100$, except for wastes. Finally, the width and the height of any waste area must be at least $w_\text{min} = 20$. This last constraint has an unusual consequence as illustrated in Figure~\ref{fig:mincuts}.
\end{itemize}

The objective is to minimize the total waste area. It differs from classical Bin Packing Problems in that the remaining part of the last bin is not counted as waste. This objective is known in the packing literature as Bin Packing with Leftovers.
It can be formulated as:
\[ \min \;\;\; nHW - Hw - \sum_{i \in \mathcal{I}} w_i h_i  \]
where $n$ is the number of bins used; $W$ and $H$ are respectively the standardized width and height of the bins; $w$ is the position of the last $1$-cut; $\mathcal{I}$ is the set of produced items; and $w_i$ and $h_i$ are respectively the width and the height of item $i \in \mathcal{I}$.

%% file: 2_definitions.tex
We use the following vocabulary: a $k$-cut is a cut performed in the $k$-th stage. Cuts separate bins or sub-plates in $k$-th level sub-plates.
For example, $1$-cuts separate the bin in several first level sub-plates. $S$ denotes a solution or a node in the search tree.

\begin{figure}
  \centering
  \begin{tikzpicture}
    \draw[fill=gray!30] (0,0) rectangle (4,3);
    \draw[fill=white] (0,0) rectangle (1,3) node[pos=.5] {$J_1$};
    \draw[fill=white] (1,0) rectangle (3,1) node[pos=.5] {$J_2$};
    \draw[fill=white] (1,1) rectangle (2,2) node[pos=.5] {$J_3$};
    \draw[fill=white] (2,1) rectangle (3.5,1.5) node[pos=.5] {$J_4$};
    \draw[dotted] (3.5,0) -- (3.5,3);
    \draw[dotted] (1,2) -- (3.5,2);
    \draw (3.5,0) node[anchor=north] {$x_1^\text{curr} = x_3^\text{curr}$};
    \draw (1,0) node[anchor=north] {$x_1^\text{prev}$};
    \draw (2,0) node[anchor=north] {$x_3^\text{prev}$};
    \draw (0,1) node[anchor=east] {$y_2^\text{prev}$};
    \draw (0,2) node[anchor=east] {$y_2^\text{curr}$};
    \draw[pattern=north west lines, pattern color=blue] (0,0) rectangle (1,3);
    \draw[pattern=north west lines, pattern color=blue] (1,0) rectangle (3.5,2);
  \end{tikzpicture}
  \caption{Last bin of a solution which does not contain all items. The area is the whole hatched part and the waste in the grey hatched part.}
  \label{fig:area}
\end{figure}
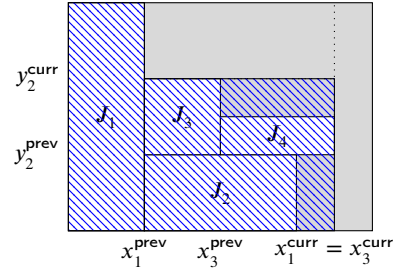

We call the last first level sub-plate, the rightmost one containing an item; the last second level sub-plate, the topmost one containing an item in the last first level sub-plate; and the last third level sub-plate the rightmost one containing an item in the last second level sub-plate.
$x_1^\text{prev}(S)$ and $x_1^\text{curr}(S)$ are the left and right coordinates of the last first level sub-plate; $y_2^\text{prev}(S)$ and $y_2^\text{curr}(S)$ are the bottom and top coordinates of the last second level sub-plate; and $x_3^\text{prev}(S)$ and $x_3^\text{curr}(S)$ are the left and right coordinates of the last third level sub-plate. Figure \ref{fig:area} presents a usage example of these definitions. We define the area and the waste of a solution $S$ as follows:

To compute $\mathrm{area}(S)$ we distinguish two cases

\begin{itemize}
\item if $S$ contains all items:
    \[ \mathrm{area}(S) = x_1^\text{curr}(S) h \]
\item and otherwise:
\begin{displaymath}
    \mathrm{area}(S) = \begin{array}{lll}
        A &+& x_1^\text{prev}(S) h \\
          &+& (x_1^\text{curr}(S) - x_1^\text{prev}(S)) y_2^\text{prev}(S) \\
          &+& (x_3^\text{curr}(S) - x_1^\text{prev}(S)) (y_2^\text{curr}(S) - y_2^\text{prev}(S)) \\
    \end{array}
\end{displaymath}
\end{itemize}

We compute the waste of a partial solution as follows:
$$ \mathrm{waste}(S) = \mathrm{area}(S) - \mathrm{item\_area}(S) $$

with $A$ the sum of the areas of all but the last bin, $h$ the height of the last bin and $\mathrm{item\_area}(S)$ the sum of the area of the items of $S$. Area and waste are illustrated in Figure~\ref{fig:area}.

%% file: 2_branching_scheme.tex
\subsection{General scheme}
\label{ssec:gen}

Two kinds of packing strategies are used in the packing literature: item-based and block-based. In item based strategies, only one item is inserted at each step, whereas in block-based strategies, multiple items are inserted. Although several researchers highlighted the benefits of block-based approaches \citep{bortfeldt_tree_2012,wei_block-based_2014,lodi_partial_2017}, we chose an item-based strategy. Two reasons support this choice. First, the problem has more constraints than classical packing problems from the literature. Thus, generating feasible solutions is already challenging and block-based approaches add even more complexity. Second, the benefits of the block-based approaches might be compensated by a more powerful tree search algorithm.

However, our strategy is not purely item-based: instead of packing one item at each step, we pack the next third level sub-plate. This comes from the observation that because only one $4$-cut is allowed in a third level sub-plate, a third level sub-plate has only five possible configurations; it may contain
\begin{enumerate}[noitemsep]
  \item exactly one item, without waste
  \item exactly one item with some waste above
  \item exactly one item with some waste below
  \item exactly two items, without waste
  \item no item, only waste
\end{enumerate}

These configurations are illustrated in Figure~\ref{fig:subplate}. The sub-plates containing $J_1$ and $J_2$ respectively follow configurations $1$ and $2$. These are the \say{standard} configurations. Placing an item on top of the sub-plate as in configuration 3 may be necessary if it would contain a defect otherwise as $J_3$. Similarly, inserting only waste (configuration $5$) may also be necessary if the region contains a defect as the sub-plate containing the second defect. We do not allow directly inserting only waste in a region containing no defects. Such sub-plate may appear in a solution, as the third-level sub-plate to the right of $J_4$ and $J_5$, but it is implicitly generated when $J_6$ is inserted. Finally, the sub-plate containing items $J_4$ and $J_5$ corresponds to configuration $4$.

\begin{figure}[]
  \centering
  \begin{tikzpicture}
    \draw[fill=gray!30] (0,0) rectangle (6,3.21);
    \draw[fill=white] (0,0) rectangle (2,1.5) node[pos=.5] {$J_1$};
    \draw[fill=white] (2,0) rectangle (3,1.25) node[pos=.5] {$J_2$};
    \draw[dotted] (3,0) -- (3,3.21);
    \draw[dotted] (1.5,1.5) -- (3,1.5);
    \draw[fill=black] (0.4,1.65) rectangle (0.6,1.75);
    \draw[fill=white] (0,1.75) rectangle (1,2.5) node[pos=.5] {$J_3$};
    \draw[fill=black] (1.3,1.9) rectangle (1.5,2.1);
    \draw[fill=white] (1.5,1.5) rectangle (2.75,2) node[pos=.5] {$J_4$};
    \draw[fill=white] (1.5,2) rectangle (2.75,2.5) node[pos=.5] {$J_5$};
    \draw[dotted] (0,2.5) -- (3,2.5);
    \draw[dotted] (1,1.5) -- (1,2.5);
    \draw[fill=white] (0,2.5) rectangle (3,3.21) node[pos=.5] {$J_6$};
    %\draw[fill=gray!60] (5,0) rectangle (6,3.21);
  \end{tikzpicture}
  \caption{Illustration of third level sub-plate possible configurations. Black rectangles are defects.}
  \label{fig:subplate}
\end{figure}

Third level sub-plates are inserted in the order they are extracted. In Figure~\ref{fig:subplate}, this follows the numbering of the items. This ensures to never violate the precedence constraints. All items are candidates if their insertion does not lead to a precedence constraint violation.

Then, a third level sub-plate can be inserted at several depths
\begin{itemize}[noitemsep]
  \item depth $0$: in a new bin
  \item depth $1$: in a new first level sub-plate to the right of the current one
  \item depth $2$: in a new second-level sub-plate above the current one
  \item depth $3$: in the current second-level sub-plate already, to the right of the last third-level sub-plate
\end{itemize}

To reduce the size of the tree, we apply some simple pruning rules:
\begin{itemize}[noitemsep]
  \item if a third-level sub-plate can be inserted in the current bin, we do not consider insertions in a new bin; and if a third level sub-plate can be inserted in the current first (resp.\ second) level sub-plate without increasing the position of its left $1$-cut (resp.\ top $2$-cut), we do not consider insertions in a new first (resp.\ second) level sub-plate;
  \item If the last insertion is an empty sub-plate at depth $d$, then the next insertion must also happen at depth $d$;
  \item If the last insertion is a $2$-item insertion at depth $d \neq 3$, then the next insertion must be at depth $3$.
\end{itemize}

With this branching scheme, item rotation and minimum and maximum distances between cuts constraints are easy to take into account.

\subsection{Pseudo-dominance rule}
\label{ssec:pseudo_dom}

In this section, we describe a more sophisticated heuristic dominance rule.
For a (partial) solution, we define its \say{front} as the polygonal chain
\begin{align*}
(
(x^\text{curr}_1, 0),
(x^\text{curr}_1, y^\text{prev}_2),
(x^\text{curr}_3, y^\text{prev}_2), \\
(x^\text{curr}_3, y^\text{curr}_2),
(x^\text{curr}_1, y^\text{curr}_2),
(x^\text{curr}_1, h)
)
\end{align*}

Figure~\ref{tikz:front} shows two examples of solution fronts.

\begin{figure}[]
  \centering
  \begin{tikzpicture}
    \draw[fill=gray!30] (0,0) rectangle (6,3.21);
    \draw[fill=white] (0,0) rectangle (2,1.5) node[pos=.5] {$J_1$};
    \draw[fill=white] (0,1.5) rectangle (2,3) node[pos=.5] {$J_2$};
    \draw[fill=white] (2,0) rectangle (3,1.25) node[pos=.5] {$J_3$};
    \draw[fill=white] (3,0) rectangle (4,1) node[pos=.5] {$J_4$};
    \draw[fill=white] (2,1.25) rectangle (2.5,2) node[pos=.5] {$J_5$};
    \draw[fill=gray!60] (4,0) rectangle (6,3.21);
    \draw[red, ultra thick] (2,3.21) -- (2,2) -- (2.5,2) -- (2.5,1.25) -- (4,1.25) -- (4,0);
  \end{tikzpicture}

  \bigskip

  \begin{tikzpicture}
    \draw[fill=gray!30] (0,0) rectangle (6,3.21);
    \draw[fill=white] (0,0) rectangle (2,1) node[pos=.5] {$J_1$};
    \draw[fill=white] (0,1) rectangle (1,2) node[pos=.5] {$J_2$};
    \draw[fill=gray!60] (2,0) rectangle (6,3.21);
    \draw[red, ultra thick] (0,3.21) -- (0,2) -- (1,2) -- (1,1) -- (2,1) -- (2,0);
  \end{tikzpicture}
  \caption{Illustration of the front of two partial solutions}
  \label{tikz:front}
\end{figure}

Then we say that solution $S_1$ dominates solution $S_2$ iff they contain the same items and the front of $S_1$ is \say{before} the front of $S_2$.
(see Figure~\ref{tikz:pseudo_dom_example}).

\begin{figure}[]
  \begin{subfigure}[t]{0.5\textwidth}
    \centering
    \begin{tikzpicture}
      \draw[fill=gray!30] (0,0) rectangle (6,3.21);
      \draw[fill=white] (0,0) rectangle (2,1) node[pos=.5] {$J_1$};
      \draw[fill=white] (0,1) rectangle (1,3) node[pos=.5] {$J_2$};
      \draw[fill=white] (1,1) rectangle (2,3) node[pos=.5] {$J_3$};
      \draw[fill=gray!60] (2,0) rectangle (6,3.21);
      \draw[red, ultra thick] (0,3.21) -- (0,3) -- (2,3) -- (2,0);
    \end{tikzpicture}
    \caption{}
  \end{subfigure}

  \medskip

  \begin{subfigure}[t]{0.5\textwidth}
    \centering
    \begin{tikzpicture}
      \draw[fill=gray!30] (0,0) rectangle (6,3.21);
      \draw[fill=white] (0,0) rectangle (2,1) node[pos=.5] {$J_1$};
      \draw[fill=white] (0,1) rectangle (1,3) node[pos=.5] {$J_2$};
      \draw[fill=white] (1,1) rectangle (3,2) node[pos=.5] {$J_3$};
      \draw[fill=gray!60] (3,0) rectangle (6,3.21);
      \draw[red, ultra thick] (0,3.21) -- (0,3) -- (3,3) -- (3,0);
    \end{tikzpicture}
    \caption{}
  \end{subfigure}

  \caption{Solution (a) dominates solution (b)}
  \label{tikz:pseudo_dom_example}
\end{figure}

If the number of possible subsets of items is small, then for a given subset, we can memorize the best front currently seen during the search and prune any new dominated node encountered. This situation occurs in instances with strong precedence constraints (\textit{i.e.}\ two chains) and this is the strategy of the DPA* algorithm presented afterward. However, for most instances, the number of possible subsets is too large and we only use the pseudo-dominance rule among the children of a node. To compensate, an additional symmetry breaking strategy is introduced.

\subsection{Symmetry breaking strategy}
\label{ssec:sym}

We designed the following symmetry breaking strategy: if they do not contain defects and can be exchanged without violating the precedence constraints, a $k$-level sub-plate is forbidden to contain an item with a smaller index than the previous $k$ level sub-plate of the same $(k-1)$-level sub-plate.

\begin{figure}[]
  \centering
  \begin{subfigure}[t]{0.5\textwidth}
    \centering
    \begin{tikzpicture}
      \draw[fill=gray!30] (0,0) rectangle (6,3.21);
      \draw[fill=white] (0,0) rectangle (1.75,2) node[pos=.5] {$J_2$};
      \draw[fill=white] (0,2) rectangle (2,3) node[pos=.5] {$J_1$};
      \draw[fill=white] (2,0) rectangle (3,3.21) node[pos=.5] {$J_3$};
      \draw[fill=gray!60] (3,0) rectangle (6,3.21);
    \end{tikzpicture}
    \caption{}
  \end{subfigure}

  \medskip

  \begin{subfigure}[t]{0.5\textwidth}
    \centering
    \begin{tikzpicture}
      \draw[fill=gray!30] (0,0) rectangle (6,3.21);
      \draw[fill=white] (0,0) rectangle (1.75,2) node[pos=.5] {$J_2$};
      \draw[fill=white] (0,2) rectangle (2,3) node[pos=.5] {$J_1$};
      \draw[fill=white] (2,0) rectangle (3,3.21) node[pos=.5] {$J_3$};
      \draw[fill=black] (1.8,0.8) rectangle (1.95,1);
      \draw[fill=gray!60] (3,0) rectangle (6,3.21);
    \end{tikzpicture}
    \caption{}
  \end{subfigure}

  \medskip

  \begin{subfigure}[t]{0.5\textwidth}
    \centering
    \begin{tikzpicture}
      \draw[fill=gray!30] (0,0) rectangle (6,3.21);
      \draw[fill=white] (0,0) rectangle (1.75,2) node[pos=.5] {$J_2$};
      \draw[fill=white] (0,2) rectangle (1,3) node[pos=.5] {$J_1$};
      \draw[fill=white] (1,2) rectangle (2,3) node[pos=.5] {$J_3$};
      \draw[fill=white] (2,0) rectangle (3,3.21) node[pos=.5] {$J_4$};
      \draw[fill=gray!60] (3,0) rectangle (6,3.21);
    \end{tikzpicture}

    $J_1$, $J_2 \to J_3$
    \caption{}
  \end{subfigure}

  \caption{Illustration of the symmetry breaking strategy: pattern (a) is forbidden because the second-level sub-plates containing $J_1$ and $J_2$ can be exchanged without a feasibility issue. However, pattern (b) is allowed because of the defect and pattern (c) is also allowed because if the second-level sub-plates are exchanged, then the precedence constraint between $J_2$ and $J_3$ is violated.}
  \label{tikz:symmetries_1}
\end{figure}
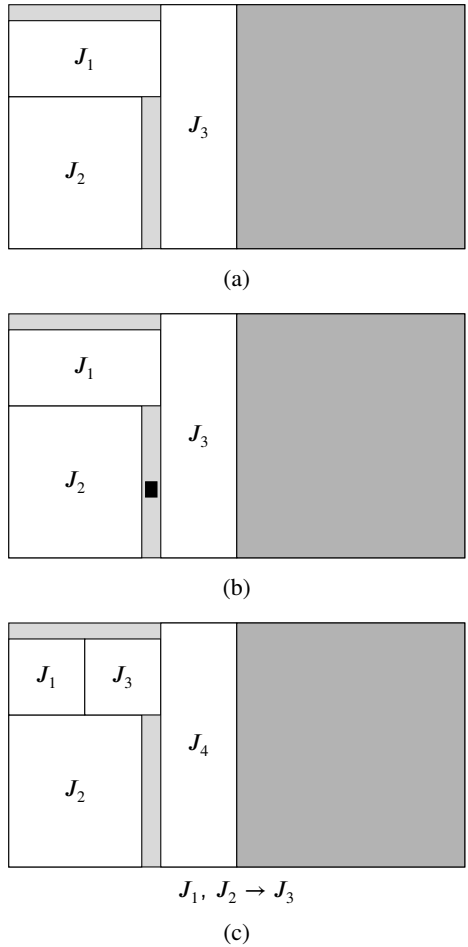

Preliminary experiments showed that applying the strategy for $k = 2$ and $k = 3$ yield the best results. The symmetry breaking strategy is illustrated in Figure~\ref{tikz:symmetries_1}.

\bigskip

It should be noted that the branching scheme is not dominant, \textit{i.e.}\ for some instances, it may not contain an optimal solution. Likewise, the pseudo-dominance rule considers that solution $S_1$ dominates solution $S_2$ whereas no optimal solution can be reached from $S_1$ but one can be from $S_2$. More details about this are given by~\cite{fontan_theoretical_2019}.

%% file: 3_tree_search.tex
\paragraph{}During our initial work on the challenge, we first explored the classical \emph{``Operations Research''} optimization algorithms (local-search, evolutionary algorithms and branch and bounds). However, it seemed difficult for us to find efficient local-search or evolutionary moves, while it felt relatively natural to design constructive methods. We implemented several classical constructive algorithms: a greedy algorithm quickly providing solutions but with limited quality; a Best First (A*) algorithm returning the ``optimal'' one (relatively to the branching scheme) on small instances; and a Depth First struggling to improve the greedy solution.

\begin{algorithm}
  \caption{A*}
  \label{A}
  fringe $\gets \{ \mathrm{root} \}$\;
  \While{$\mathrm{fringe} \neq \emptyset$ and $\mathrm{time} < \mathrm{time limit}$}{
    $n \gets extractBest(\mathrm{fringe})$\;
    $\mathrm{fringe} \gets \mathrm{fringe} \setminus \{ n \}$\;
    \ForAll{ $v \in neighbours(n)$ }{
      $\mathrm{fringe} \gets \mathrm{fringe} \cup \{ v \}$\;
    }
  }
\end{algorithm}

We remind the pseudo-code of the A* algorithm in Algorithm~\ref{A}.
At each iteration, the \say{best} node is extracted from the fringe and its children are added to the fringe. As written above, our implementation of A* was able to find the optimal solutions on very small instances but was quickly running out of memory larger ones because of the size of the fringe. Therefore, we decided to heuristically prune nodes to bound the required memory. This ``heuristic'' algorithm performed beyond expectations and provided excellent solutions. However, it depended on the amount of memory allowed for the fringe. If this parameter is too small, the search ends quickly and not benefit from the remaining available time. If too big, the search takes more time and does not provide any solution within the time limit. To get rid of this parameter, we chose to use a restart strategy where we geometrically increase the allowed memory at each restart. The new parameter to calibrate becomes the growth factor, but we found that any value between $1.25$ and $3$ provided similar results. This simple approach provided good solutions. 

Motivated by this simple but yet efficient algorithm, we investigated other anytime tree search algorithms such as beam search \citep{ow1988filtered} and beam stack search \citep{zhou2005beam}. We implemented and compared them on the challenge problem. To our surprise, they did not perform as well as the previously described approach. To the best of our knowledge, this approach has not been used in the Operations Research literature before. We describe it in more detail in the next section.

\subsection{Memory Bounded A* (MBA*)}

A* is known to minimize the cost estimate on nodes it opens.
However, it suffers from a large memory requirement since it has to store a large number of nodes in the fringe. 
We propose a simple but yet powerful heuristic variant of A* that cuts less promising nodes if the size of the fringe goes over a parameter $D$.
We call this tree search algorithm \emph{Memory Bounded A* (MBA*)}. 
If $D = 1$, it generalizes a greedy algorithm and if $D = \infty$, it generalizes A*.

The pseudo-code of MBA* is given in Algorithm~\ref{MBA}.
Only lines 8 to 11 are added compared to the A* algorithm presented in Algorithm~\ref{A}.
MBA* starts with a fringe containing only the root node (line 1). At each iteration, the \say{best} node is extracted from the queue (lines 3 and 4) and its children are added to the queue (lines 5 to 7).
If the size of the queue goes over $D$, the \say{worst} nodes are discarded (lines 8 to 11).

\begin{algorithm}
  \caption{Memory Bounded A* (MBA*)\label{MBA}}
  fringe $\gets \{ \mathrm{root} \}$\;
  \While{$\mathrm{fringe} \neq \emptyset$ and $\mathrm{time} < \mathrm{time limit}$}{
    $n \gets extractBest(\mathrm{fringe})$\;
    $\mathrm{fringe} \gets \mathrm{fringe} \setminus \{ n \}$\;
    \ForAll{ $v \in neighbours(n)$ }{
      $\mathrm{fringe} \gets \mathrm{fringe} \cup \{ v \}$\;
    }
    \While { $|\mathrm{fringe}| > D$ }{
      $n \gets extractWorst(\mathrm{fringe})$\;
      $\mathrm{fringe} \gets \mathrm{fringe} \setminus \{ n \}$\;
    }
  }
\end{algorithm}

\subsection{Guide functions}

\begin{figure*}
  \centering
  \begin{subfigure}[b]{\textwidth}
    \includegraphics[width=0.5\textwidth]{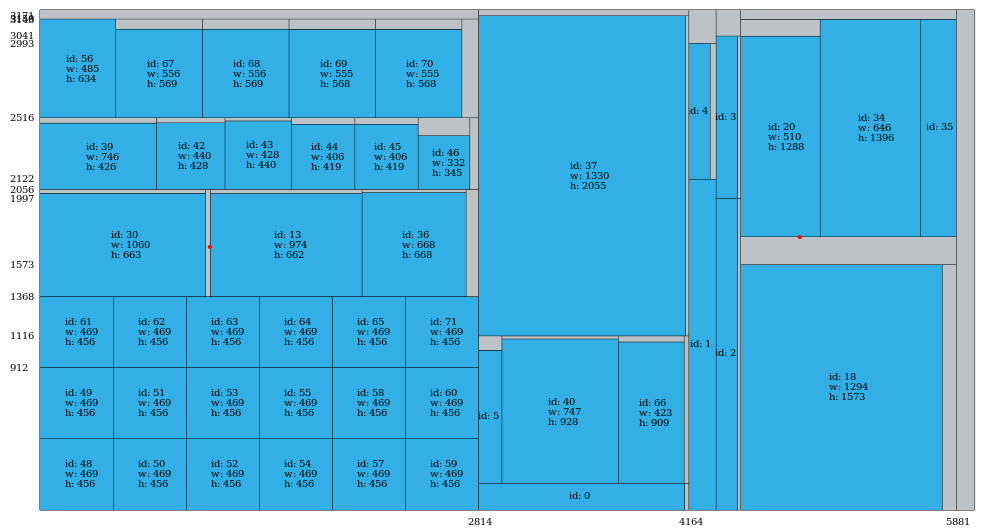}
    \includegraphics[width=0.5\textwidth]{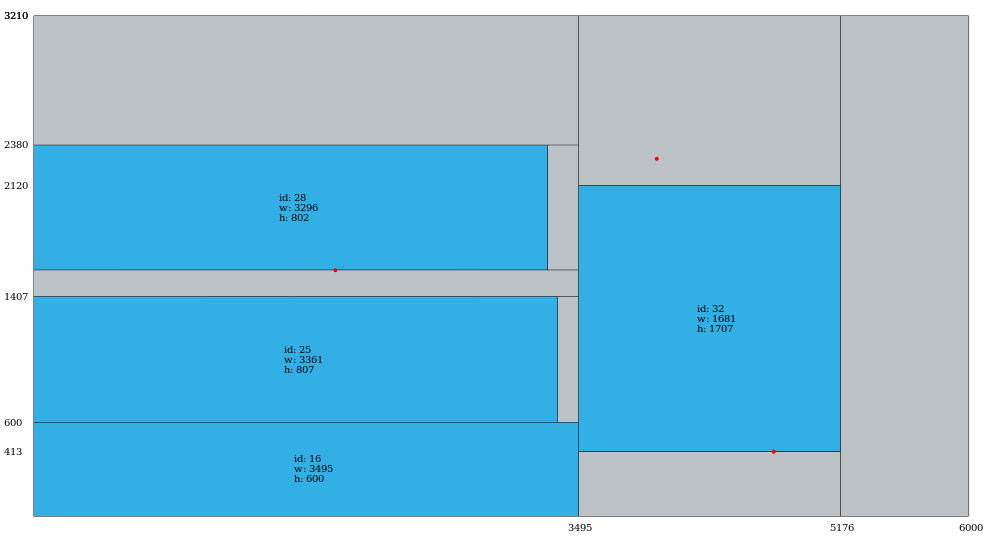}
    \caption{waste only guide -- first and last plate}
    \label{fig:naiveguide1}
  \end{subfigure}

  \medskip

  \begin{subfigure}[b]{\textwidth}
    \includegraphics[width=0.5\textwidth]{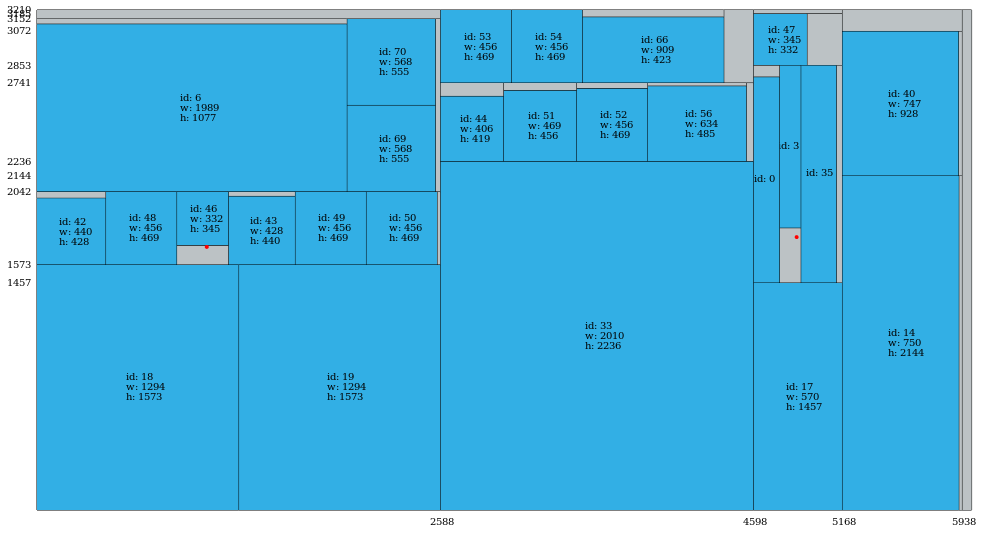}
    \includegraphics[width=0.5\textwidth]{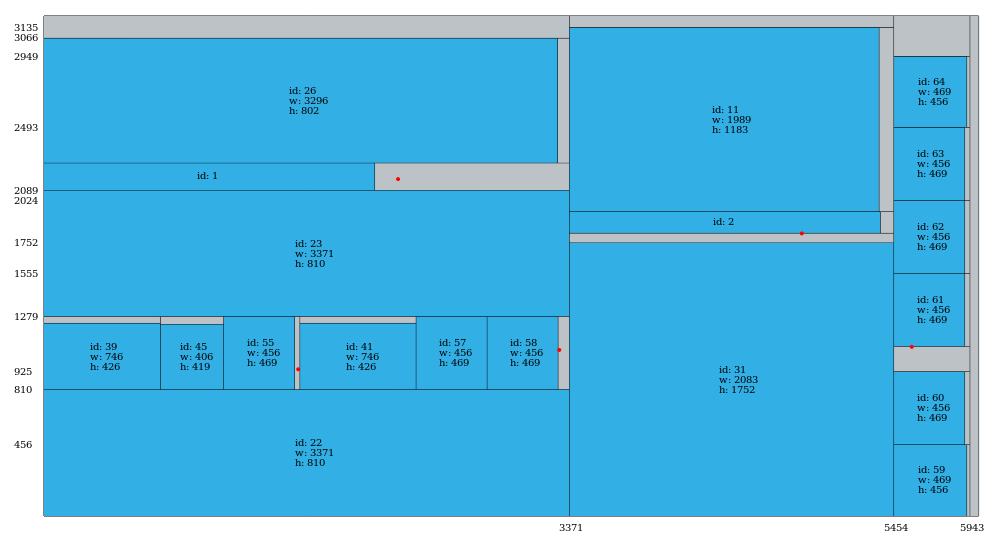}
    \caption{waste / average size guide -- first and last plate}
    \label{fig:newguide1}
  \end{subfigure}
  \caption{\ref{fig:naiveguide1} shows a solution obtained using the waste as guide function. Notice that at the beginning of the solution, small items are omnipresent whereas in later plates, only large items are present, thus globally generates more waste.
  \ref{fig:newguide1} shows the effect of the guide biased by the item average size on a solution of the same instance. We observe that small and large items are better mixed and significantly less waste is generated at the end of the solution.}
  \label{fig:naiveguide}
\end{figure*}

Tree search methods are dependent on well-crafted guide functions which define the meaning of \say{best} and \say{worst} nodes.
Using a lower bound is common in the tree search literature. Indeed, if the objective is to prove optimality, using a lower bound as a guide function will minimize the number of opened nodes.
Therefore, we first tried this approach and used the waste as a guide function. We noticed that the resulting solutions packed small items on the first plates and big items on the last ones, thus generating little waste in the beginning but a lot in the end. Globally, the solution quality was not satisfactory as illustrated in Figure~\ref{fig:naiveguide1}.

Taking this into account, we designed new guides to balance the cost of inserting small items at the beginning of the solutions:
\begin{description}
  \item[waste percentage (= waste / area):] \hfill\\
    compared to waste only, the waste has less impact if the solution contains larger items.
  \item[waste percentage / average area of packed items:] \hfill\\
    this guide function directly adds a reward to solutions containing large items.
\end{description}

The benefit of these guides is illustrated in Figure~\ref{fig:newguide1}.

\subsection{DPA*: solving instances with strong precedence constraints}

Three instances of the challenge contain only 2 precedence chains. If we denote by $n_1$ (resp.\ $n_2$) the length of the first (resp.\ second) chain, then the number of possible subsets of items packed in a partial solution of the branching scheme becomes $n_1n_2$. Since the number of items in an instance is less than $700$ (this information was given in the challenge description), it becomes possible to store the non-dominated fronts encountered for each possible subset without overcoming the memory limitation, compare the front of each opened node with the non-dominated fronts from all the previously encountered nodes and prune the dominated ones.

Therefore, we developed a dedicated algorithm for these instances named Dynamic Programming A* (DPA*). DPA* is an A* algorithm implementing the scheme described in the previous paragraph. DPA* does not bound the size of the queue as MBA* does, and uses the waste as a guide function. Therefore, if it terminates, it returns the \say{optimal} solution (relatively to the branching scheme and the pseudo-dominance rule).

We may note that for a given subset of items, there could be an exponential number of non-dominated fronts, which could degrade DPA* performances. This is at least not an issue for the concerned instances from the challenge.

\subsection{Global algorithm}

%The configuration of the computer used to evaluate submissions of the competition was
%Linux OS machine (Ubuntu 18.04) with 4 CPU (8 Threads) with 3.6GHz -Intel Xeon W-2123 Processor, 48GB of RAM and a graphic card of 5GB.

For the competition, we distinguished the case where the instance has two chains or less. 
In this case, we run DPA*. 

If it has strictly more than two chains, we do not use DPA* since it would overcome the memory limitation. 
Since the processor used to evaluate the participant submissions had $4$ physical cores, we run 4 threads, each one running a restarting MBA* with a given growth factor and a given guide function.
Each MBA* is initially executed with a fringe maximal size of $2$, and each time one terminates, it is restarted with a maximal fringe size multiplied by its growth factor.
If the growth factor is 2, the maximal size doubles at each iteration. 
All the threads share the information of the best solution found.
If one finds a better solution, the others take advantage of it to perform more cuts and globally perform better together than alone.
The threads run the following algorithms:
\begin{itemize}
    \item MBA*, waste percentage guide, growth factor 1.33
    \item MBA*, waste percentage guide, growth factor 1.5
    \item MBA*, waste percentage / average size guide, growth factor 1.33
    \item MBA*, waste percentage / average size guide, growth factor 1.5
\end{itemize}

%% file: 4_num_results.tex
In this section, we first evaluate the contribution of the components we described in the previous sections in the main algorithm. Then, we show the benefits of using DPA* on instances with only two precedence chains. Finally, we provide computational results with the challenge setting.
Instances generally have between 300 and 600 items and 10 to 15 chains.
They are available online\footnote{\url{https://www.roadef.org/challenge/2018/en/instances.php}}.

\subsection{Contribution of the components}

In this section, computational experiments have been performed on a personal computer with an Intel(R) Core(TM) i5-3470 CPU @ 3.20GHz with 16GB RAM.

We consider the $12$ possible combinations of the components we designed, namely MBA* to be compared with an Iterative Beam Search \cite{zhang1998complete}; with or without the symmetry breaking strategy; and with waste (w), waste percentage (p), or waste percentage/average size (a) guide. We run each pair of instance-algorithm for $100$ seconds.

In Table~\ref{tab:alg_comp}, we show a summary of the performances of each variant. The goal is to help us select the best combinations to use. MBA* with the symmetry breaking strategy and guided by the waste percentage (\emph{MBA*+sym+p}) and MBA* with the symmetry breaking strategy and guided by the waste percentage / average size (\emph{MBA*+sym+a}) clearly outperform all other combinations. That is why these are the two combinations that we use. Since the processor used to evaluate participant submissions has 4 physical cores, we dedicate two threads for each combination with different growth factor (1.33 and 1.5) to add some robustness.

\begin{table}[]
    \centering
\begin{tabular}{l|r|r}
  \toprule
Combination	& \# best	& \# only best	\\
\midrule
BS+no-sym+w	& 2	& 0	\\
BS+no-sym+p	& 2	& 0	\\
BS+no-sym+a	& 2	& 0	\\
BS+sym+w	& 5	& 1	\\
BS+sym+p	& 9	& 4	\\
BS+sym+a	& 9	& 4	\\
MBA*+no-sym+w	& 2	& 0	\\
MBA*+no-sym+p	& 3	& 0	\\
MBA*+no-sym+a	& 4	& 0	\\
MBA*+sym+w	& 3	& 0	\\
MBA*+sym+p	& 23	& 17	\\
MBA*+sym+a	& 22	& 18	\\
\bottomrule
\end{tabular}
    \caption{Comparison over all possible algorithms using the proposed algorithmic components. ``\# best'' indicate the number of times the algorithm was able to find the best solution on given instances compared to the 11 other algorithms. ``\# only best'' indicates the number of times the algorithm was the only one to find the best solution.}
    \label{tab:alg_comp}
\end{table}

\begin{table*}[h]
  \centering
  \begin{tabular}{l|r|r||r|r||r|r|r}
    \toprule
    Instance	& best IBS	& best MBA*	& best no sym	& best with sym	& best w	& best p	& best a	\\
    \midrule
    A1	& \textbf{425 486}	& \textbf{425 486}	& \textbf{425 486}	& \textbf{425 486}	& \textbf{425 486}	& \textbf{425 486}	& \textbf{425 486}	\\
    A2	& 10 514 609	& \textbf{9 676 799}	& 10 537 079	& \textbf{9 676 799}	& 10 659 059	& 10 418 309	& \textbf{9 676 799}	\\
    A3	& \textbf{2 651 880}	& \textbf{2 651 880}	& 3 441 540	& \textbf{2 651 880}	& 3 056 340	& \textbf{2 651 880}	& \textbf{2 651 880}	\\
    A4	& 3 242 520	& \textbf{3 220 050}	& 3 306 720	& \textbf{3 220 050}	& 3 505 740	& \textbf{3 220 050}	& 3 306 720	\\
    A5	& \textbf{3 033 273}	& 3 566 133	& 4 856 553	& \textbf{3 033 273}	& 3 736 263	& \textbf{3 033 273}	& 3 566 133	\\
    A6	& \textbf{3 225 930}	& 3 572 610	& 3 652 860	& \textbf{3 225 930}	& 3 800 520	& \textbf{3 225 930}	& 3 460 260	\\
    A7	& 5 063 280	& \textbf{4 800 060}	& 5 194 890	& \textbf{4 800 060}	& 5 933 190	& \textbf{4 800 060}	& 4 938 090	\\
    A8	& \textbf{9 187 874}	& 10 077 044	& 12 218 114	& \textbf{9 187 874}	& 12 568 004	& 10 077 044	& \textbf{9 187 874}	\\
    A9	& \textbf{2 930 706}	& 2 985 276	& 3 550 236	& \textbf{2 930 706}	& 3 929 016	& \textbf{2 930 706}	& 2 985 276	\\
    A10	& 4 097 221	& \textbf{4 084 381}	& 5 272 081	& \textbf{4 084 381}	& 4 797 001	& \textbf{4 084 381}	& 4 122 901	\\
    A11	& \textbf{4 718 449}	& 4 978 459	& 6 076 279	& \textbf{4 718 449}	& 6 711 859	& 5 251 309	& \textbf{4 718 449}	\\
    A12	& \textbf{2 050 084}	& 2 245 894	& 2 342 194	& \textbf{2 050 084}	& \textbf{2 050 084}	& 2 104 654	& 2 194 534	\\
    A13	& 15 096 453	& \textbf{12 133 623}	& 14 865 333	& \textbf{12 133 623}	& 15 099 663	& 12 197 823	& \textbf{12 133 623}	\\
    A14	& 14 363 778	& \textbf{12 097 518}	& 14 793 918	& \textbf{12 097 518}	& 14 363 778	& \textbf{12 097 518}	& 13 490 658	\\
    A15	& 15 277 961	& \textbf{13 185 041}	& 15 168 821	& \textbf{13 185 041}	& 16 029 101	& \textbf{13 185 041}	& 15 014 741	\\
    A16	& \textbf{3 380 333}	& \textbf{3 380 333}	& \textbf{3 380 333}	& \textbf{3 380 333}	& \textbf{3 380 333}	& \textbf{3 380 333}	& \textbf{3 380 333}	\\
    A17	& \textbf{3 617 251}	& \textbf{3 617 251}	& \textbf{3 617 251}	& \textbf{3 617 251}	& \textbf{3 617 251}	& \textbf{3 617 251}	& \textbf{3 617 251}	\\
    A18	& 5 898 468	& \textbf{5 535 738}	& 5 763 648	& \textbf{5 535 738}	& 7 737 798	& 5 596 728	& \textbf{5 535 738}	\\
    A19	& \textbf{3 323 744}	& 3 654 374	& 4 187 234	& \textbf{3 323 744}	& 4 620 584	& \textbf{3 323 744}	& 3 965 744	\\
    A20	& \textbf{1 467 925}	& \textbf{1 467 925}	& 1 493 605	& \textbf{1 467 925}	& \textbf{1 467 925}	& \textbf{1 467 925}	& \textbf{1 467 925}	\\
    \midrule
    B1	& 4 173 228	& \textbf{3 633 948}	& 4 150 758	& \textbf{3 633 948}	& 4 012 728	& 4 324 098	& \textbf{3 633 948}	\\
    B2	& 15 715 685	& \textbf{15 359 375}	& 18 466 655	& \textbf{15 359 375}	& 20 155 115	& \textbf{15 359 375}	& 15 715 685	\\
    B3	& 32 668 193	& \textbf{21 253 433}	& 23 365 613	& \textbf{21 253 433}	& 41 315 933	& 24 890 363	& \textbf{21 253 433}	\\
    B4	& 8 885 365	& \textbf{8 862 895}	& 11 238 295	& \textbf{8 862 895}	& 9 222 415	& \textbf{8 862 895}	& 8 920 675	\\
    B5	& 92 433 185	& \textbf{88 590 815}	& \textbf{88 590 815}	& \textbf{88 590 815}	& 103 719 545	& \textbf{88 590 815}	& \textbf{88 590 815}	\\
    B6	& \textbf{13 371 637}	& 13 480 777	& 15 653 947	& \textbf{13 371 637}	& 17 509 327	& 14 113 147	& \textbf{13 371 637}	\\
    B7	& 14 576 799	& \textbf{11 434 209}	& 12 801 669	& \textbf{11 434 209}	& 14 319 999	& \textbf{11 434 209}	& 12 801 669	\\
    B8	& 24 490 999	& \textbf{19 512 289}	& 24 121 849	& \textbf{19 512 289}	& 24 490 999	& \textbf{19 512 289}	& 20 048 359	\\
    B9	& 20 511 607	& \textbf{20 046 157}	& 25 085 857	& \textbf{20 046 157}	& 46 721 257	& 39 071 827	& \textbf{20 046 157}	\\
    B10	& 28 012 013	& \textbf{27 344 333}	& 29 225 393	& \textbf{27 344 333}	& 35 815 523	& 31 055 093	& \textbf{27 344 333}	\\
    B11	& 38 143 250	& \textbf{29 113 520}	& 34 175 690	& \textbf{29 113 520}	& 41 523 380	& 32 589 950	& \textbf{29 113 520}	\\
    B12	& 18 122 077	& \textbf{16 086 937}	& 19 929 307	& \textbf{16 086 937}	& 18 122 077	& \textbf{16 086 937}	& 16 314 847	\\
    B13	& 31 138 545	& \textbf{29 674 785}	& 33 716 175	& \textbf{29 674 785}	& 31 138 545	& 32 213 895	& \textbf{29 674 785}	\\
    B14	& 10 482 820	& \textbf{10 043 050}	& 11 619 160	& \textbf{10 043 050}	& 12 046 090	& 10 434 670	& \textbf{10 043 050}	\\
    B15	& 41 533 241	& \textbf{28 372 241}	& 34 143 821	& \textbf{28 372 241}	& 41 533 241	& \textbf{28 372 241}	& 31 466 681	\\
    \midrule
    X1	& 21 022 877	& \textbf{17 299 277}	& 17 970 167	& \textbf{17 299 277}	& 29 911 367	& 17 803 247	& \textbf{17 299 277}	\\
    X2	& 11 459 837	& \textbf{8 583 677}	& 8 923 937	& \textbf{8 583 677}	& 9 318 767	& 9 206 417	& \textbf{8 583 677}	\\
    X3	& 9 424 756	& \textbf{8 712 136}	& 9 842 056	& \textbf{8 712 136}	& 9 578 836	& \textbf{8 712 136}	& 8 927 206	\\
    X4	& 19 035 422	& \textbf{15 976 292}	& 19 305 062	& \textbf{15 976 292}	& 19 035 422	& \textbf{15 976 292}	& 16 772 372	\\
    X5	& \textbf{5 383 037}	& 5 620 577	& 7 029 767	& \textbf{5 383 037}	& 6 728 027	& 5 623 787	& \textbf{5 383 037}	\\
    X6	& 14 443 523	& \textbf{12 167 633}	& 14 488 463	& \textbf{12 167 633}	& 14 443 523	& 13 024 703	& \textbf{12 167 633}	\\
    X7	& 30 327 120	& \textbf{26 170 170}	& 27 146 010	& \textbf{26 170 170}	& 31 328 640	& 29 161 890	& \textbf{26 170 170}	\\
    X8	& 27 693 711	& \textbf{27 109 491}	& 27 693 711	& \textbf{27 109 491}	& 27 693 711	& 27 494 691	& \textbf{27 109 491}	\\
    X9	& 33 431 655	& \textbf{23 599 425}	& 33 919 575	& \textbf{23 599 425}	& 33 370 665	& \textbf{23 599 425}	& 26 716 335	\\
    X10	& 23 400 722	& \textbf{19 901 822}	& 23 522 702	& \textbf{19 901 822}	& 23 673 572	& 23 975 312	& \textbf{19 901 822}	\\
    X11	& 14 349 972	& \textbf{14 247 252}	& 16 102 632	& \textbf{14 247 252}	& 14 349 972	& \textbf{14 247 252}	& 14 921 352	\\
    X12	& 14 775 805	& \textbf{12 422 875}	& 14 576 785	& \textbf{12 422 875}	& 14 775 805	& \textbf{12 422 875}	& 12 589 795	\\
    X13	& 19 208 322	& \textbf{14 624 442}	& 18 900 162	& \textbf{14 624 442}	& 20 007 612	& 16 271 172	& \textbf{14 624 442}	\\
    X14	& 11 075 552	& \textbf{9 730 562}	& 11 916 572	& \textbf{9 730 562}	& 11 004 932	& \textbf{9 730 562}	& 10 128 602	\\
    X15	& 16 301 394	& \textbf{13 540 794}	& 17 261 184	& \textbf{13 540 794}	& 16 461 894	& 13 990 194	& \textbf{13 540 794}	\\
    \midrule
    total waste	& 779 159 574	& 679 871 064	& 779 027 964	& 676 914 654	& 870 817 914	& 725 241 204	& 693 016 014	\\
    \midrule
    nb best	& 14	& 41	& 4	& 50	& 5	& 27	& 28	\\
    \midrule
    nb only best	& 9	& 36	& 0	& 46	& 1	& 21	& 22	\\
    \bottomrule
  \end{tabular}
  \caption{Analysis of the contribution of each introduced algorithmic component}
  \label{tab:comp_contrib}
\end{table*}

Table~\ref{tab:comp_contrib} presents an analysis of the contribution of each component individually. Each column corresponds to the best result per instance obtained by a subset of algorithms that uses a given component. For instance \emph{best IBS} corresponds to a subset of algorithms using Iterative Beam Search, thus excluding MBA* (6 algorithms). MBA* variants outperform the Iterative Beam Search variants (producing 12\% less waste). It finds 41/50 best solutions and 36 best solutions that the Beam Search variants were not able to obtain. The algorithms using the symmetry breaking strategy clearly produce better results than the one without (13\% less waste and 46 best solutions not attainable by the variants without the symmetry breaking strategy) showing that integrating state-space reductions can greatly benefit to anytime tree search algorithms and probably even to constructive meta-heuristics. Finally, as expected, the waste (lower bound) guide provides the worst results among the 3 considered guides (16\% more waste than the waste percentage guide and 20\% more waste than the waste percentage / average size guide and only 5 best solutions on 50 instances). However, the waste percentage guide and the waste percentage / average size guide provided similar results (with a slight advantage on the latest as it produces 5\% less waste and finds one best solution more). These results are interesting as they show that both guides are complementary. Indeed, they produce 21 (resp. 22) best solutions where they are the only one to obtain them. Thus it is worth using both.

\subsection{DPA*}

Table~\ref{tab:dpastar} shows the benefits of using DPA* on instances with only two precedence chains. There is one such instance in each dataset, but the one in dataset A is trivial to solve, therefore we only consider instances B5 and X8.
Unlike MBA*, DPA* is not anytime and terminates long before the $3600$ seconds time limit.
Furthermore, the solutions it returns are significantly better and are even the best-known solutions for both instances.

\begin{table}[htpb]
  \centering
  \begin{tabular}{l|c|c}
    \toprule
    Instance & \begin{tabular}{c}MBA* \\ (4 threads, 3600s)\end{tabular} & DPA* \\
    \midrule
    B5 & $88\,590\,815$ & $72\,155\,615$ ($2.06$s) \\
    X8 & $24\,875\,331$ & $22\,265\,601$ ($59.12$s) \\
    \bottomrule
  \end{tabular}
  \caption{DPA* vs MBA*}
  \label{tab:dpastar}
\end{table}

\subsection{Final results}

Table~\ref{fig:comp} sums up the challenge final results. Computational experiments have been performed on a computer with an Intel Core i7-4790 CPU @ 3.60 GHz $\times$ 8 processor with 31.3 Go of RAM.
This configuration is similar to the one of the challenge.
Since the challenge, a few adjustments have been made. Therefore, the results presented here slightly differ from the results obtained during the final phase.
Compared to the challenge version, the current version performs better:
the total waste on dataset B and X decreases from $493,600,549$ for the challenge version to $469,910,749$ for the current one.
Columns \emph{Final phase best 180s} and \emph{Final phase best 3600s} contain the best solutions found during the final phase.
Results annotated with a star indicate that this solution was found by our algorithm during the final phase of the challenge.
The \emph{Best known} column contains the best solution up to our knowledge. 
They may have been found during the development of the algorithm, with execution times exceeding $3600$ seconds or by other teams.
Finally, even if it is not indicated in the table, on most of the instances, if the algorithm is run longer, for example, $2$ hours, the solution will still be improved.

\begin{table*}
  \small
  \centering
  \begin{tabular}{l|c||r|r||r|r|r}
    \toprule
    Instance &
    Comments        &
    \begin{tabular}{c}Final phase \\ best 180s \end{tabular} &
    \begin{tabular}{c}MBA*/DPA* \\ 180s \end{tabular} &
    \begin{tabular}{c}Final phase \\ best 3600s \end{tabular} &
    \begin{tabular}{c}MBA*/DPA* \\ 3600s \end{tabular} &
    Best known \\
    \midrule
    A1   & Trivial        & -          & 425 486    &       -      & 425 486    & 425 486    \\
    A2   & No prec        & -          & 9 506 669  &       -      & 4 383 509  & 4 383 509  \\
    A3   &                & -          & 2 651 880  &       -      & 2 651 880  & 2 651 880  \\
    A4   &                & -          & 3 024 240  &       -      & 2 924 730  & 2 924 730  \\
    A5   &                & -          & 2 924 730  &       -      & 3 283 653  & 3 017 223  \\
    A6   &                & -          & 3 389 640  &       -      & 3 225 930  & 3 188 646  \\
    A7   &                & -          & 4 703 760  &       -      & 4 334 610  & 3 920 520  \\
    A8   &                & -          & 9 691 844  &       -      & 8 378 954  & 8 378 954  \\
    A9   &                & -          & 2 664 276  &       -      & 2 664 276  & 2 664 276  \\
    A10  &                & -          & 4 084 381  &       -      & 4 084 381  & 4 084 381  \\
    A11  &                & -          & 4 660 669  &       -      & 4 622 149  & 4 358 929  \\
    A12  &                & -          & 2 056 504  &       -      & 1 879 954  & 1 879 954  \\
    A13  &                & -          & 10 226 883 &       -      & 9 440 433  & 9 331 293  \\
    A14  &                & -          & 11 686 638 &       -      & 10 383 378 & 10 383 378 \\
    A15  &                & -          & 12 918 611 &       -      & 11 108 171 & 10 828 901 \\
    A16  & Trivial        & -          & 3 380 333  &       -      & 3 380 333  & 3 380 333  \\
    A17  & 2 chains       & -          & 3 617 251  &       -      & 3 617 251  & 3 617 251  \\
    A18  &                & -          & 5 596 728  &       -      & 4 983 618  & 4 983 618  \\
    A19  &                & -          & 3 654 374  &       -      & 3 323 744  & 3 323 744  \\
    A20  & Trivial        & -          & 1 467 925  &       -      & 1 467 925  & 1 467 925  \\
    \midrule
    B1   & No prec        & 3 232 698  & 3 765 558  & \textbf{*2 661 318}    & 3 136 398  & 2 661 318  \\
    B2   &                & *15 635 435 & 14 312 915 & *13 674 125 & 13 398 065 & 11 931 095 \\
    B3   &                & 20 540 813 & 19 786 463 & 18 191 093   & 17 093 273 & 15 786 803 \\
    B4   &                & *8 269 045  & 8 323 615  & *8 269 045    & 7 973 725  & 7 315 675  \\
    B5   &     2 chains   & 72 155 615 & 72 155 615 & \textbf{72 155 615}   & \textbf{72 155 615} & 72 155 615 \\
    B6   &                & *12 116 527 & 12 488 887 & *11 195 257   & 11 089 327 & 10 800 427 \\
    B7   & No prec        & 9 601 299  & 9 177 579 & *8 355 819 & 7 678 509 & 6 628 839 \\
    B8   &                & *17 865 559 & 17 152 939 & 16 067 959 & 15 840 049 & 14 398 759 \\
    B9   &                & 18 502 147 & 19 969 117 & 17 484 577   & 17 474 947 & 16 495 897 \\
    B10  &                & 26 012 183 & 26 904 563 & \textbf{*21 951 533}   & 23 065 403 & 21 951 533 \\
    B11  &                & 25 251 890 & 27 312 710 & 22 584 380   & 23 820 230 & 20 626 280 \\
    B12  &                & *15 868 657 & 13 734 007 & *13 958 707   & 13 120 897 & 12 514 207 \\
    B13  &                & *28 349 055 & 27 360 375 & *24 471 375   & 23 078 235 & 22 657 725 \\
    B14  &                & *9 346 480  & 9 442 780  & *8 656 330    & 8 377 060  & 8 023 960 \\
    B15  &                & *27 794 441 & 24 568 391 & *24 517 031   & 23 088 581 & 22 619 921 \\
    \midrule
    X1   &                & *15 508 097 & 15 302 657 & *14 127 797 & 14 127 797 & 13 720 127 \\
    X2   & No prec        & 6 034 937  & 6 083 087 & *5 434 667 & 4 879 337 & 4 795 877  \\
    X3   &                & *8 285 206  & 7 649 626  & *7 473 076    & 7 180 966  & 6 837 496  \\
    X4   &                & 12 182 072 & 15 488 372 & \textbf{11 405 252}   & 13 366 562 & 11 405 252  \\
    X5   &                & 5 081 297  & 4 988 207 & 4 712 147 & 4 715 357 & 4 522 757  \\
    X6   &                & 12 565 673 & 11 031 293 & *10 363 613   & 9 496 913 & 9 365 303 \\
    X7   &                & *22 443 360 & 22 876 710 & 21 127 260   & 21 191 460 & 20 568 720 \\
    X8   &    2 chains    & *24 788 661 & 22 265 601 & *24 788 661   & \textbf{22 265 601} & 22 265 601 \\
    X9   &                & *22 251 225 & 22 312 215 & 20 167 935   & 20 479 305 & 20 039 535 \\
    X10  &                & *20 110 472 & 18 778 322 & *17 824 952   & 17 186 162 & 16 865 162 \\
    X11  &                & *13 489 692 & 12 802 752 & *12 417 552 & 11 676 042 & 11 011 572 \\
    X12  &                & *11 963 845 & 12 358 675 & *10 583 545   & 10 503 295 & 10 246 495 \\
    X13  &                & 15 950 172 & 14 345 172 & *13 533 042   & 13 125 372 & 12 130 272 \\
    X14  &                & *8 889 542  & 8 591 012 & *8 013 212 & 7 644 062 & 7 422 572 \\
    X15  &                & 13 990 194 & 13 710 924 & 11 682 204   & 11 682 204 & 10 882 914 \\
    \bottomrule
  \end{tabular}
  \caption{Computational experiments comparing the proposed approach compared to other contestants}
  \label{fig:comp}
\end{table*}

%% file: 5_conclusion.tex
In this article, we presented a new anytime tree search algorithm called MBA* for the 2018 ROADEF/EURO challenge glass cutting problem. It performs successive iterations, restarting when its heuristic search tree exploration is completed. During the first iterations, it performs aggressive prunings and behaves like a greedy algorithm. As iterations go, the algorithm performs fewer heuristic prunings, and thus gets access to better solutions (at the cost of an increase of the computation time of each iteration).
If enough time and memory are available, the algorithm ends up performing an iteration with no heuristic pruning, finding the best solution regarding the branching scheme.

We proposed two guides (waste percentage, and waste percentage / average item size). These guides can find significantly better solutions than using a lower bound (the waste), which is what is usually used in branch and bounds. We also presented a symmetry breaking strategy and showed that it significantly improves the quality of the solutions returned by the algorithm. 

Also, we designed another algorithm, DPA*, for instances with only two precedence chains. This algorithm returns the best-knowns on these instances within short times.

\bigskip

This result shows that anytime tree search algorithms from the AI planning community, and branch and bounds from the Operations Research community can benefit from each other, leading to algorithms competitive with classical meta-heuristics. 
We believe that the representation of anytime tree search algorithms in the Operations Research literature does not reflect the benefits of applying such methods on complex optimization problems. To the best of our knowledge, many of them remain unexplored such as Beam Stack Search \citep{zhou2005beam}, SMA* \citep{russell1992cient} or Anytime Column Search \citep{cohen2018anytime}.

Motivated by the success of MBA* on this glass cutting application, {\color{red}fontan2020} already adapted it for classical guillotine cutting problems from the literature and showed that even on more fundamental problems, it is still competitive with the other dedicated algorithms from the literature.